\documentclass{article}

\usepackage{microtype}
\usepackage{graphicx}
\usepackage{subfigure}
\usepackage{booktabs} 
\usepackage{algorithm}
\usepackage{algorithmic}
\usepackage{multirow}
\usepackage{subcaption}
\usepackage{graphicx}
\usepackage{array}

\usepackage{hyperref}

\usepackage[accepted]{icml2025}

\usepackage{amsmath}
\usepackage{amssymb}
\usepackage{mathtools}
\usepackage{amsthm}

\usepackage[capitalize,noabbrev]{cleveref}

\theoremstyle{plain}

\theoremstyle{definition}

\theoremstyle{remark}

\newcommand{\RETURN}{\textbf{Return }}

\usepackage[textsize=tiny]{todonotes}

\icmltitlerunning{FedEM: A Privacy-Preserving Framework for Concurrent Utility Preservation in Federated Learning}

\begin{document}

\twocolumn[
\icmltitle{FedEM: A Privacy-Preserving Framework for Concurrent Utility Preservation in Federated Learning}




\icmlsetsymbol{equal}{*}

\begin{icmlauthorlist}
\icmlauthor{Mingcong Xu}{sch}
\icmlauthor{Xiaojin Zhang}{sch}
\icmlauthor{Wei Chen}{sch}
\icmlauthor{Hai Jin}{sch}
\end{icmlauthorlist}



\icmlkeywords{Federated Learning,privacy protecton,privacy-utility tradeoff}

\vskip 0.3in
]


\begin{abstract}

Federated Learning (FL) enables collaborative training of models across distributed clients without sharing local data, addressing privacy concerns in decentralized systems. However, the gradient-sharing process exposes private data to potential leakage, compromising FL's privacy guarantees in real-world applications.  To address this issue, we propose Federated Error Minimization (FedEM), a novel algorithm that incorporates controlled perturbations through adaptive noise injection.  This mechanism effectively mitigates gradient leakage attacks while maintaining model performance.  Experimental results on benchmark datasets demonstrate that FedEM significantly reduces privacy risks and preserves model accuracy, achieving a robust balance between privacy protection and utility preservation.

\end{abstract}

\section{Introduction}

Federated learning has emerged as a promising paradigm for collaborative machine learning, enabling multiple clients to jointly train a global model without directly sharing their local data~\cite{mcmahan2017communication, li2024comprehensive}. By preserving data decentralization, FL addresses privacy concerns while leveraging the diverse data distributions across clients. However, despite its advantages, FL is still vulnerable to privacy threats. Adversaries can exploit weaknesses in gradient-sharing techniques, which makes it challenging to design reliable and privacy-preserving FL systems.

Existing attack techniques, such as membership inference~\cite{shokri2017membership}, property inference~\cite{melis2019exploiting}, and gradient leakage attacks (GLA)~\cite{zhu2019deep}, can compromise client privacy in FL environments. Among these, GLAs have drawn significant attention because they exploit shared gradients to recover the original training data, potentially revealing sensitive information about clients. These threats highlight the urgent need for effective privacy protection mechanisms in FL.

Several methods have been proposed to mitigate privacy risks in FL. Encryption-based techniques~\cite{xu2019verifynet} offer robust privacy guarantees but introduce substantial computational and communication overhead, limiting scalability in resource-constrained environments. Differential privacy (DP) approaches, such as Centralized DP (CDP)\cite{geyer2017differentially} and Local DP (LDP)\cite{sun2020ldp}, provide alternative solutions. However, these methods often degrade model performance due to the noise they introduce, particularly in LDP settings where noise is directly added to gradients. Achieving an optimal balance between privacy and utility remains a persistent challenge in FL research.

\begin{figure}[htbp]
\centering
\includegraphics[width=1.0\linewidth]{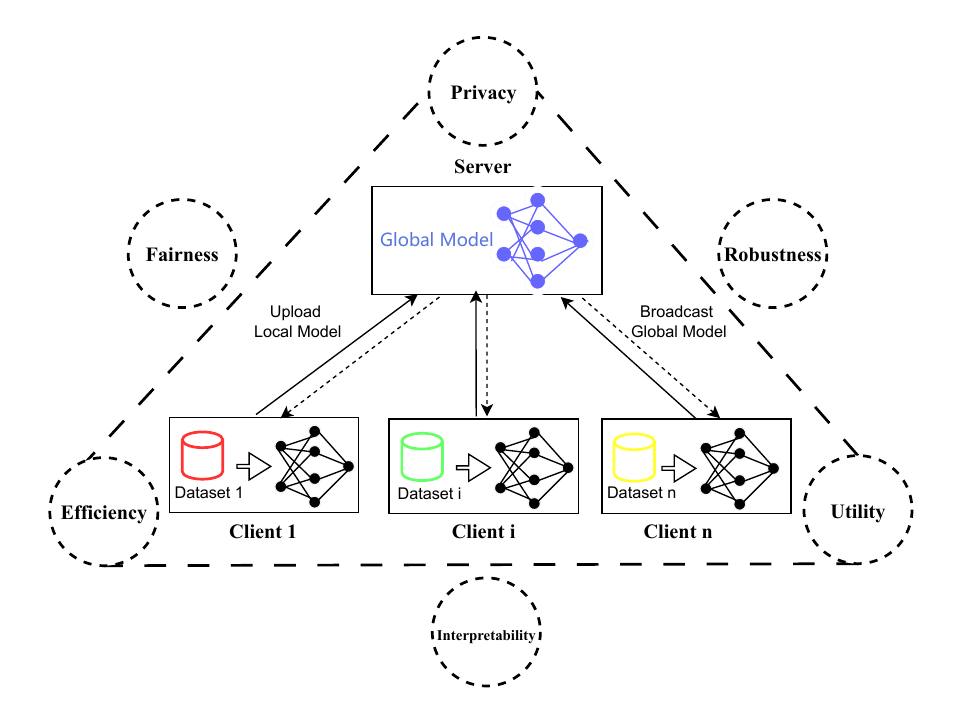}
\caption{Multi-objective trade-offs in Federated Learning.}
\label{fig:tradeoff}
\end{figure}

This trade-off between privacy, utility, and other objectives such as efficiency, robustness, and fairness is central to the design of FL systems. These interdependencies are illustrated in Figure~\ref{fig:tradeoff}, which highlights the multi-objective nature of FL optimization. Effective FL solutions must carefully navigate these trade-offs to achieve a balanced system.

This trade-off between privacy, utility, and other objectives such as efficiency, robustness, and fairness is fundamental to the design of FL systems. These interdependencies are depicted in Figure~\ref{fig:tradeoff}, which illustrates the multi-objective nature of FL optimization. Effective FL solutions must carefully navigate these trade-offs to achieve a balanced system. In this work, we draw inspiration from data poisoning techniques and introduce a novel algorithm, FedEM, aimed at enhancing privacy while minimizing performance degradation. Unlike traditional LDP methods, which inject noise into gradients, FedEM incorporates controlled perturbations directly into the client data. These perturbations are carefully crafted to reduce the risk of data reconstruction while maintaining model utility. By reformulating the FL optimization objective to account for perturbation constraints, FedEM achieves a more favorable balance between utility and privacy protection.

We conduct comprehensive experiments to evaluate FedEM’s effectiveness. The results demonstrate that FedEM outperforms established privacy-preserving techniques, in safeguarding privacy while maintaining competitive model performance.

The main contributions of this work are summarized as follows:

\begin{itemize} 
\item We highlight the limitations of current privacy-preserving techniques in FL and stress the need for novel approaches to mitigate gradient leakage attacks. 
\item We propose FedEM, a data perturbation-based algorithm that effectively defends against gradient leakage attacks while preserving model utility, achieving an efficient privacy-utility trade-off. 
\item We evaluate FedEM on multiple datasets and compare it with state-of-the-art methods, demonstrating its superiority in providing robust privacy protection in FL systems. 
\end{itemize}

\section{Related Works}

\subsection{Attack Methods in Federated Learning}
Federated learning decentralizes data among clients to ensure privacy, but significant privacy risks remain. Adversaries can exploit specific vulnerabilities to extract private information from client data. Existing attack methods in FL are generally categorized into three main types: membership inference attacks~\cite{salem2018ml, nasr2019comprehensive, song2021systematic, hui2021practical}, property inference attacks~\cite{ganju2018property, melis2019exploiting, wang2023attrleaks}, and gradient leakage attacks~\cite{zhu2019deep, zhao2020idlg}.

Membership inference attacks leverage the fact that models typically perform better on training data than on unseen data, enabling adversaries to determine whether a particular data point is part of the training set. Fredrikson et al.~\cite{fredrikson2015model} showed that model predictions can reveal sensitive information about training data, laying the groundwork for this research. Shokri et al.~\cite{shokri2017membership} formalized these attacks, allowing adversaries to identify whether specific data samples are part of the training set. Building on this, property inference attacks were introduced to infer private characteristics of the training data.

Gradient leakage attacks, first proposed by Zhu et al.~\cite{zhu2019deep}, highlighted how the gradient-sharing mechanism in FL could lead to significant privacy leakage. These attacks aim to reconstruct original training samples, such as images, from shared gradient updates. This approach, known as Deep Leakage from Gradients (DLG), solves the following optimization problem:

\begin{equation}\label{attack_optim}
\min_{{x}}\min_{{y}}{\parallel \nabla_{\theta}\mathcal{L}(x,y)-g\parallel }
\end{equation}

Here, \(\nabla_{\theta}\mathcal{L}(x, y)\) is the gradient computed using the reconstructed input \(x\) and label \(y\), and \(g\) represents the true gradients. By minimizing the discrepancy between the gradients from reconstructed samples and shared gradients, the attack reconstructs data samples that closely resemble the originals. Subsequent work, such as Zhao et al.~\cite{zhao2020idlg}, improved DLG for high-resolution images, while recent GLA techniques incorporate regularization terms to enhance reconstruction performance for large-batch and high-resolution inputs~\cite{geiping2020inverting, yin2021see, yue2023gradient}.

Besides optimization-based methods, analytical approaches have been explored. Zhu et al.~\cite{zhu2020r}, Chen et al.~\cite{chen2021understanding}, and Lu et al.~\cite{lu2022april} utilized gradient properties to reconstruct client data directly. Unlike optimization techniques, these methods rely on explicit gradient analysis rather than iterative optimization.

Both optimization-based and analytical GLA methods generally assume a semi-honest server, which passively observes shared parameters without interfering in the FL process. However, recent studies have introduced malicious servers and clients~\cite{fowl2021robbing, fowl2022decepticons, boenisch2023curious}, who actively manipulate shared parameters or model weights, leading to more sophisticated attacks.

\subsection{Privacy-Preserving Mechanisms in Federated Learning}
The two main categories of privacy protection strategies used for federated learning are methods based on DP and those utilizing encryption techniques.

Encryption-based methods, including homomorphic encryption (HE) and secure multi-party computation (SMPC), safeguard data privacy. HE allows encrypted computations without decryption~\cite{aono2017privacy, madi2021secure}, but the substantial computational and communication overhead limits its scalability~\cite{bonawitz2017practical}. SMPC, using secret sharing, partitions data among participants~\cite{xu2019verifynet, zhao2022pvd}, providing strong privacy guarantees without relying on a trusted server. However, its high computational complexity makes it less practical in resource-constrained environments.

DP-based methods are also widely adopted in federated learning and can be classified into centralized differential privacy and local differential privacy~\cite{jiang2024fed}. Centralized DP~\cite{geyer2017differentially, miao2022against} assumes the server is a trusted entity and adds DP noise during the aggregation process to mitigate membership inference and attribute inference attacks. Despite its effectiveness in these scenarios, CDP offers limited protection against gradient leakage attacks. In contrast, local differential privacy  introduces DP noise to gradients before they are uploaded by the clients~\cite{sun2020ldp, liu2020fedsel, kim2021federated, wang2023ppefl}, effectively addressing gradient leakage vulnerabilities. However, this approach significantly degrades model performance due to the direct application of noise to gradient data. To address this limitation, researchers have proposed incorporating shuffling mechanisms into LDP~\cite{girgis2021shuffled}, which reduce the required noise magnitude and thereby achieve a better equilibrium between model utility and privacy.

Recent studies have also provided theoretical analyses of the privacy-utility trade-off in federated learning. Specifically, ~\cite{zhang2023probably, zhang2023game} conducted a theoretical examination of this trade-off, offering insights into how privacy and utility interact in the context of FL. Additionally, ~\cite{zhang2023theoretically, zhang2024meta} explored advanced perturbation strategies, aiming to find an optimal balance between utility performance and privacy protection from a practical application perspective.

\subsection{Unlearnable Examples}
Adversarial training is a technique that introduces perturbations to the data in order to enhance model robustness. ~\cite{huang2021unlearnable} introduced the concept of unlearnable algorithms, where carefully crafted perturbations are applied to the data to prevent unauthorized machine learning models from effectively extracting meaningful information. This method aims to both protect data privacy and improve model robustness. The core optimization objective is:

\begin{equation} \begin{aligned} \min_{\theta }\max_{\delta } & \quad \mathbb{E}_{(x, y) \sim T}[\mathcal{L}(f(x + \delta; \theta), y)], \ \text{s.t.} & \quad ||\delta| | \leq \epsilon, \end{aligned} \end{equation}

where $\theta$ represents the model parameter, $\delta$ is the adversarial perturbation added to data $x$, $\epsilon$ constrains the perturbation magnitude, $T$ denotes the data distribution, and $\mathcal{L}$ represents model loss. The inner maximization $\max_{\delta}$ identifies perturbations that degrade model performance by maximizing classification loss, while the outer minimization $\min_{\theta}$ optimizes model parameters to mitigate this effect. This ensures the data becomes unlearnable by unauthorized models while retaining usability for legitimate tasks.

A related method, known as the error minimization attack, leverages adversarial training to poison the training process by inducing classification errors through perturbations\cite{zheng2020evaluating}. Its optimization objective is:

\begin{equation} \begin{aligned} \min_{\theta }\min_{\delta } & \quad \mathbb{E}_{(x, y) \sim T}[\mathcal{L}(f(x + \delta; \theta), y)], \ \text{s.t.} & \quad  | | \delta| |  \leq \epsilon, \end{aligned} \end{equation}

where the inner minimization $\min_{\delta}$ seeks perturbations that subtly disrupt the training process, and the outer minimization $\min_{\theta}$ adjusts model parameters to align with the poisoned data. Unlike unlearnable algorithms, which focus on protecting privacy, this method explicitly aims to degrade model performance by corrupting the training process. To protect privacy in our work, we employ the error minimization technique.

\section{The FedEM algorithm for privacy protection}

\subsection{Federated Learning}
We consider a federated learning setup involving $K$ clients, each holding its own private dataset, denoted as $T_k$. The goal of this system is to collaboratively train a globally optimal model through multiple rounds of model distribution and aggregation, without directly sharing local data. Specifically, the optimization objective is formulated as:
\begin{equation}
\min_{\theta} \sum_{k=1}^{K} \frac{m_k}{m} \mathcal{L}_k(\theta),
\end{equation}
where $\theta$ represents the global model's parameters, and $\mathcal{L}_k(\theta)$ is the objective function for client $k$. The term $m_k = |D_k|$ refers to the number of data points available to client $k$, while $m = \sum_{k=1}^{K} m_k$ is the total number of data points across all clients. 

To solve this optimization problem, several classical algorithms, such as FeSGD and FedAVG\cite{mcmahan2017communication}, have been proposed. The general procedure of these algorithms can be summarized as follows: 
\begin{enumerate}
    \item \textbf{Global Model Distribution:} At the beginning of round $t$, the server distributes the global model parameters $\theta^{(t-1)}$ to all clients.  
    \item \textbf{Local Model Update:} Upon receiving the model, all clients optimize their own loss function $\mathcal{L}_k(\theta)$ using the private dataset $D_k$ and a specified local optimization algorithm (e.g., SGD). This results in the local model parameters $\theta_k^{(t)}$ for round $t$, or other corresponding gradient information.  
    \item \textbf{Global Model Aggregation:} After updating local models, all clients send their updates back to the server. The server aggregates the updates using a specific strategy, to compute the new global model parameters $\theta^{(t)}$. 
\end{enumerate}

This process continues iteratively until the global model meets the convergence criteria or reaches a predefined maximum number of training rounds.

\subsection{Attack Model}
When considering security threats in federated systems, we assume the server acts as a semi-honest attacker. Specifically, it utilizes the gradients transmitted by the clients to carry out gradient inversion attacks, attempting to reconstruct local datasets from the shared parameters. We further assume that there are no malicious clients or servers within the system. Regarding the attacker's knowledge, we assume the server is aware of the optimization techniques and procedures used by the clients, allowing it to perform data reconstruction more effectively. The attacker's optimization objective is described in Equation \ref{attack_optim}.

\subsection{FedEM}

Existing privacy-preserving algorithms often have notable limitations. For example, LDP methods, while offering effective privacy protection, typically cause significant degradation in model performance. In contrast, Central Differential Privacy and encryption-based techniques (e.g., homomorphic encryption) are inadequate in defending against gradient reconstruction attacks, such as DLG. To address these challenges, our goal is to design an efficient privacy-preserving algorithm that not only protects clients' local data from reconstruction attacks by the server but also strives to preserve model performance.

To achieve this, we move beyond the conventional approach of adding noise to client-transmitted parameters, as used in standard LDP methods. Instead, we propose a novel mechanism, FedEM, which introduces perturbations directly to clients' local data. By strategically "poisoning" the data, FedEM effectively defends against gradient reconstruction attacks while carefully controlling the magnitude of perturbations to minimize their impact on model performance.

With the introduction of data perturbation, let $\theta$ represent the global model parameters, and let $\delta_k$ denote the local perturbation vector for the $k$-th client, constrained by norm $\rho_u^{\text{min}}$ and $\rho_u^{\text{max}}$. The input features $x_k$ and corresponding labels $y_k$ are sampled from the local data distribution $T_k$, and the predictive model $f_\theta$ minimizes the loss function $L$ applied to the perturbed data. The term $t(x_k + \delta_k)$ represents the normalization operation. The optimization objective in federated learning is reformulated as follows:

\begin{equation}
\begin{aligned}
\min_{\theta}\min_{\delta_1, \delta_2, \dots, \delta_k} \quad & \sum_{k=1}^K \frac{m_k}{m} \mathbb{E}_{(x_k,y_k) \sim T_k} \left[ L\left( f_\theta\left( t(x_k + \delta_k) \right), y_k \right) \right] \\\\
\text{s.t.} \quad & {\rho_u^{\text{min}}\leq\|\delta_k\| \leq \rho_u^{\text{max}}}.
\end{aligned}
\end{equation}

Although the optimization problem shares similarities with error minimization  attacks in its structure, the optimization objectives are fundamentally different. While EM attacks aim to poison the data to disrupt model training, we draw inspiration from this concept and apply data poisoning for privacy protection in federated learning. Specifically, we introduce a controlled data perturbation mechanism to defend against GLAs. Our goal is to balance privacy and performance while respecting the perturbation constraints.

To solve this reformulated optimization problem, FedEM operates as follows: Each client computes and updates the perturbation vector $\delta_k$ based on its local data, ensuring the perturbation norm remains within the specified bound. The perturbation is carefully designed to minimize the information leakage from local data, while still allowing the model to learn useful patterns. After training their local models with the perturbed data, clients upload the updated parameters to the server. The server aggregates these parameters, updates the global model, and broadcasts it back to the clients. This process repeats until convergence. The complete algorithm is presented in Algorithm \ref{alg}.

\begin{figure}[htbp]
    \centering
    \includegraphics[width=1.0\linewidth]{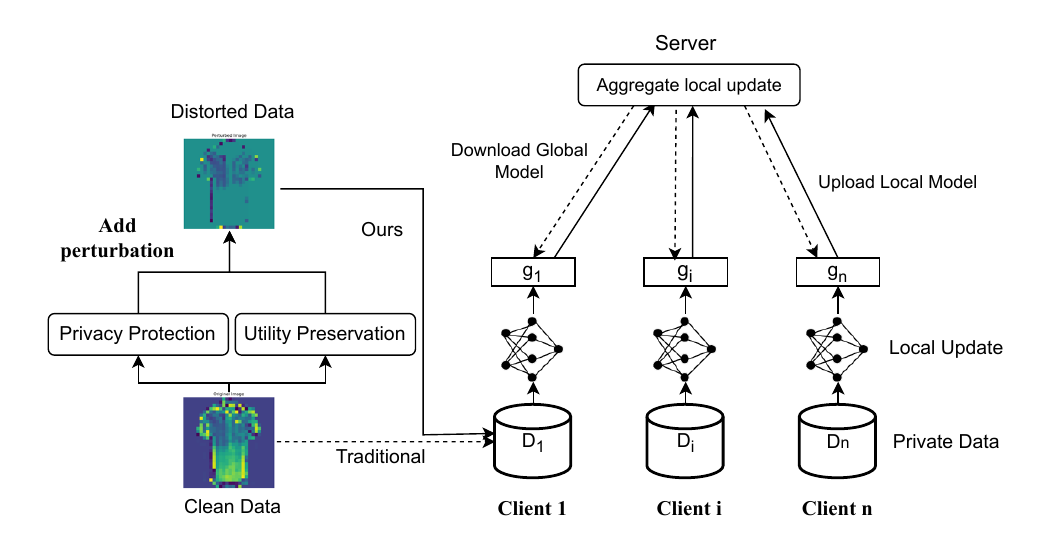}
    \caption{Frame diagram of FedFM}
    \label{fig:frame}
\end{figure}

In summary, FedEM introduces controlled perturbations at the local data level to defend against gradient reconstruction attacks while minimizing the impact on model performance. The structure of the algorithm is shown in Figure \ref{fig:frame}.

\begin{algorithm}
\caption{FedEM}
\label{alg}
\begin{algorithmic}[1]
\REQUIRE Training datasets $D_k$ (held by each client $k$), local data distribution $T_k$,  global model parameters $\theta$, defensive perturbation generator model parameters $\theta_{u}$, number of global training rounds $M$, perturbation generation parameters $\rho_{max}$, $\rho_{min}$, $\alpha_u$, normalized transformation distribution $T$, number of perturbation iterations $N$.

\ENSURE Trained global model parameters $\theta$.

\STATE Server initializes the global model parameters $\theta$.

\FOR{each global training round $t = 1$ to $M$}
    \STATE A selection of clients $C_t$  are chosen by the server and server randomly initializes the defensive perturbation $\delta_1, \delta_2, \dots, \delta_k$.
    \STATE Server broadcasts the current global model  $\theta$ and initial perturbation $\delta_1, \delta_2, \dots, \delta_k$ to all selected clients $C_t$.

    \FOR{each client $k \in C_t$ }
        \STATE Randomly choose a subset of training data $(x_k, y_k)$ from the local data distribution $T_k$, and initialize $\theta_u \leftarrow \theta$.
        \FOR{each iteration $n = 1$ to $N$}
            \STATE Update $\delta_k$: 
            \[
            \delta_k = \delta_k - \alpha_u \cdot \text{sign}(\nabla_{\delta_k} \mathcal{L}_k(f_{\theta_u}(t(x_k + \delta_k), y_k))),
            \]
            \[
            \delta_k = \text{Proj}_{\rho_{min}\leq\|\delta_k \| \leq \rho_{max}} (\delta_k).
            \]
            \STATE Update $\theta_u$ via gradient descent: 
            \[
            \theta_u = \theta_u - \eta \cdot \nabla_{\theta_u} \mathcal{L}_k(f_{\theta_u}(t(x_k + \delta_k), y_k)).
            \]
        \ENDFOR
        \STATE Generate perturbed data $x_{\text{perturbed}} = t(x_k + \delta_k)$ and compute the local gradient $g_k = \nabla_\theta {L}(f_\theta(x_{\text{perturbed}}, y_k))$.
        \STATE Upload $g_k$ to the server.
    \ENDFOR

    \STATE Server aggregates the gradients: 
    \[
    g_{\text{global}} = \frac{1}{|C_t|} \sum_{k \in C_t} g_k,
    \]
    and updates the global model: 
    \[
    \theta = \theta - \eta \cdot g_{\text{global}}.
    \]
\ENDFOR

\RETURN Trained global model parameters $\theta$.
\end{algorithmic}
\end{algorithm}

\section{Experiments}

\subsection{Experimental Settings}

\subsubsection{Datasets, attack model and baseline algorithms}

We experimented on three popular federated learning datasets to assess the FedFM algorithm's effectiveness: MNIST\cite{deng2012mnist}, FashionMNIST\cite{xiao2017fashion}, and CIFAR-10\cite{krizhevsky2009learning}. For the client-side attacks, we adopted Deep Leakage from Gradients as the primary attack method on the server side. As baselines, we selected standard Local Differential Privacy algorithms\cite{zhao2020local}, including Gaussian DP and Laplace DP. To ensure the validity and reliability of the experiments, we designed comparisons under two controlled conditions: (1) when the perturbation radius of different methods was approximately the same, and (2) when the data protection level or achieved utility was comparable, to compare the methods from another dimension.

\cite{zhang2023theoretically} also proposed an efficient privacy-preserving method. However, due to differences in the perturbation norm form between FedEM and this method, we did not include it in the baseline comparison.
Detailed description for the baseline algorithms and the experimental setup is provided in Appendix\ref{APP B} for further reference.

\begin{table*}[htbp]
\caption{Comparison of FedSGD and FedEM.}
\label{FedEM-SGD-tab}
\vskip 0.15in
\begin{center}
\begin{small}
\begin{sc}
\begin{tabular}{lccccccc}
\toprule
\textbf{Dataset} & \textbf{Methods} & \textbf{Test-Acc} & \textbf{Val-Acc} & \textbf{Test-MSE} & \textbf{Fea MSE} & \textbf{SSIM} & \textbf{PSNR} \\
\midrule
\multirow{2}{*}{MNIST} 
   & FedSGD & \textbf{0.9784} & \textbf{0.9754} & 0.9324 & 2.0676 & 0.1573 & 11.387 \\
   & FedEM(ours) & 0.9723 & 0.9684 & \textbf{1.1800} & \textbf{7.7600} & \textbf{0.1131} & \textbf{9.8471} \\  
\midrule
\multirow{2}{*}{FMNIST} 
   & FedSGD & \textbf{0.8622} & \textbf{0.8542} & 0.8064 & 0.8206 & 0.2453 & 10.588 \\
   & FedEM(ours) & 0.8499 & 0.8452 & \textbf{0.8213} & \textbf{1.7213} & \textbf{0.1454} & \textbf{10.032} \\ 
\midrule
\multirow{2}{*}{CIFAR-10}
    & FedSGD & \textbf{0.3003} & \textbf{0.2945} & \textbf{1.9111} & \textbf{14.366} & 0.0175 & \textbf{9.3946} \\
    & FedEM(ours) & 0.2554 & 0.2407 & 1.7646 & 6.5318 & \textbf{0.0170} & 9.6216 \\
\bottomrule
\end{tabular}
\end{sc}
\end{small}
\end{center}
\vskip -0.1in
\end{table*}

\subsubsection{FL settings}\label{settings}
In this sub-section, we present the FL settings adopted in our experiments. By default, the FL system consists of four clients.The global training procedure is executed for 50 rounds when using the MNIST and FMNIST datasets, and for 100 rounds when working with the CIFAR-10 dataset. The perturbation generation and local gradient optimization are both carried out using a learning rate of 0.01. We apply stochastic gradient descent (SGD) for both processes. In our experiments, the radius of the perturbation added to the data by FedFM is initialized to 8. Additionally, we set the early stopping patience to 30 epochs.

In each global round, the system follows the standard FedSGD training method \cite{mcmahan2017communication}, with the exception of the perturbation addition rules. Specifically, during each round, clients perform a single local training epoch and send their gradients directly to the server without additional local updates. We chose FedSGD because it enhances the effectiveness of DLG attacks, making the results more pronounced and easier to observe and compare.

\begin{figure*}[!htbp]
    \centering
    \includegraphics[width=\textwidth]{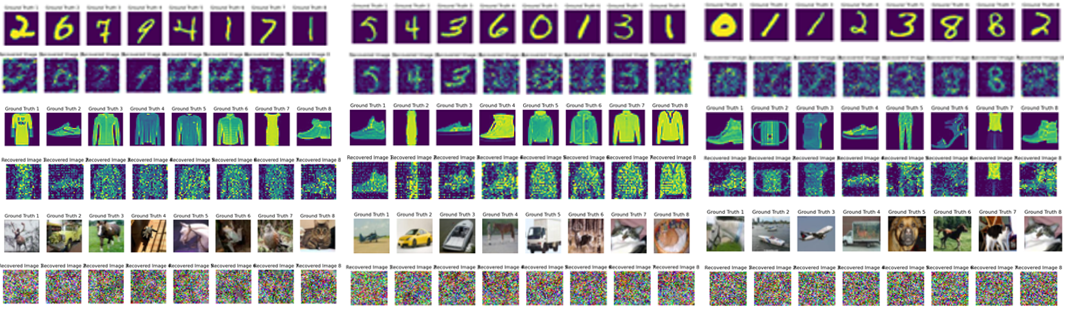} 
    \caption{DLG Reconstruction Images. From top to bottom, the first, third, and fifth rows are the original images from the MNIST, FMNIST, and CIFAR-10 datasets, respectively. The second, fourth, and sixth rows show the reconstructed images. From left to right, the algorithms are FedEM, FedSGD, and LDP.}
    \label{fig:res}
\end{figure*}

\subsubsection{EVALUATION METRICS}
In this study, we use test accuracy and validation accuracy as the primary metrics for performance evaluation. To assess privacy protection, we employ three evaluation criteria: mean squared error (MSE), structural similarity index measure (SSIM) \cite{wang2004image}, and peak signal-to-noise ratio (PSNR). These metrics compare the reconstructed images with the original ones.

\subsection{Main Results}

\begin{table*}[t]
\caption{Performance metrics across datasets and methods.}
\label{performance-table}
\vskip 0.15in
\begin{center}
\begin{small}
\begin{sc}
\begin{tabular}{lccccccc}
\toprule
Dataset & Method & Val Acc & Test Acc & Test MSE & Fea MSE & SSIM & PSNR \\
\midrule
\multirow{4}{*}{MNIST} 
    & DP-Gas & 0.7924 & 0.7947 & 0.6284 & 0.0284 & 0.4703 & 13.421 \\
    & DP-Lap & 0.6890 & 0.6796 & 0.2418 & 0.0120 & 0.6536 & 9.7231 \\
    & DP-clip & 0.9640 & 0.9578 & \textbf{1.2227} & 2.2886 & 0.1163 & \textbf{9.0826} \\
    & FedEM(ours) & \textbf{0.9723} & \textbf{0.9684} & 1.1800 & \textbf{7.7600} & \textbf{0.1131} & 9.8471 \\
\midrule
\multirow{4}{*}{FMNIST} 
    & DP-Gas & 0.7054 & 0.7048 & 0.1919 & 0.0022 & 0.6170 & 16.757 \\
    & DP-Lap & 0.6812 & 0.6810 & 0.1666 & 0.0069 & 0.5922 & 17.229 \\
    & DP-clip & 0.8461 & 0.8395 & 0.5111 & 1.6929 & 0.3667 & 14.051 \\
    & FedEM(ours) & \textbf{0.8499} & \textbf{0.8452} & \textbf{0.8213} & \textbf{1.7213} & \textbf{0.1454} & \textbf{10.032} \\
\midrule
\multirow{3}{*}{CIFAR-10} 
    & DP-Gas & 0.1490 & 0.1280 & 1.7560 & 9.5935 & \textbf{0.0142} & 9.6831 \\
    & DP-Lap & 0.1436 & 0.1381 & \textbf{1.8409} & \textbf{11.655} & 0.0169 & \textbf{9.4631} \\
    & FedEM(ours) & \textbf{0.2554} & \textbf{0.2407} & 1.7646 & 6.5318 & 0.0170 & 9.6219 \\
\bottomrule
\end{tabular}
\end{sc}
\end{small}
\end{center}
\vskip -0.1in
\end{table*}

\begin{figure*}[!htbp]
    \centering
    \includegraphics[width=\textwidth]{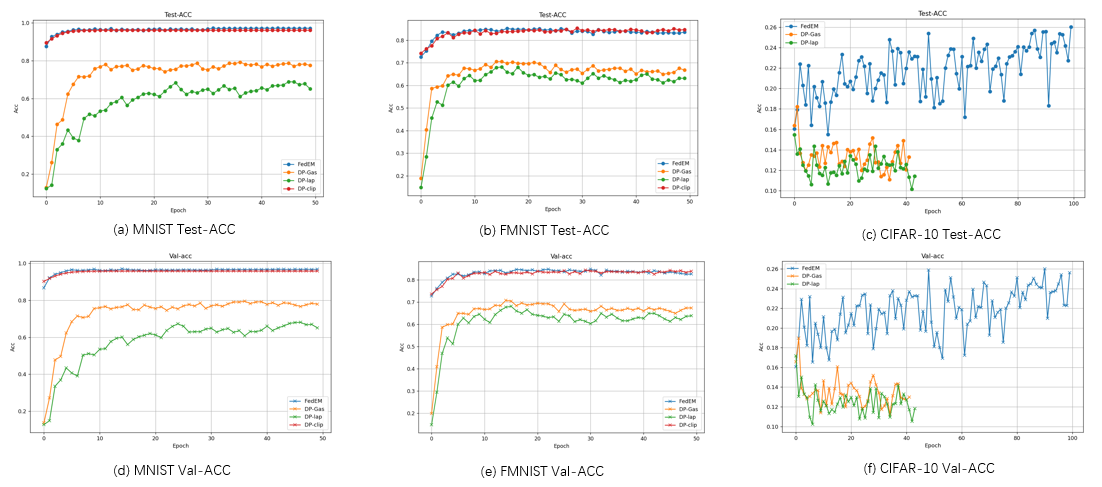} 
    \caption{Performance Evaluation of Privacy-Preserving Techniques on MNIST, FMNIST, and CIFAR-10: Test and Validation Accuracy Comparison.}
    \label{fig:Acc}
\end{figure*}

FedEM effectively mitigates privacy risks while minimizing the impact on model performance. We compare FedEM with the state-of-the-art method for defending against gradient leakage attacks, LDP. The results show that FedEM outperforms LDP in terms of model performance, while offering the same level of privacy protection. Additionally, when both methods achieve similar performance, FedEM provides significantly stronger privacy protection. The metrics for FedEM (our method) compared to FedSGD (without privacy protection) are presented in Table \ref{FedEM-SGD-tab}, while the comparison between FedEM and other privacy-preserving methods is shown in Table \ref{performance-table}. Accuracy curves for different methods are displayed in Figure~\ref{fig:Acc}, and DLG reconstructed images for various methods can be found in Figure~\ref{fig:res}.

As shown in Table~\ref{FedEM-SGD-tab} and Table~\ref{performance-table}, on the MNIST dataset, when the same level of perturbation is applied, FedEM outperforms all other methods in terms of privacy protection, with only a slight performance decrease compared to FedSGD. To further demonstrate FedEM's effectiveness, we relaxed the restriction of using the same perturbation range and applied clipping to achieve a similar privacy protection level with LDP. Under these conditions, FedEM still leads in terms of performance. This emphasizes FedEM’s robustness in preserving model utility, even when privacy protection constraints are relaxed.

On FMNIST, with the same perturbation ranges, FedEM not only offers excellent privacy protection but also incurs minimal performance loss. The results on FMNIST further validate FedEM's capability to balance privacy and performance. Even when clipping is applied to make LDP and FedEM performance comparable, FedEM continues to significantly outperform LDP in privacy protection.

For CIFAR-10, the color properties of the images reduce the impact of perturbations, and the privacy protection effectiveness of different methods is similar. However, it is important to note that while privacy protection is not drastically improved over other methods, the real strength of FedEM is evident in its ability to minimize the performance loss while still offering strong privacy guarantees. Nevertheless, the privacy protection performance of FedEM does not significantly outperform the FedSGD method, which seems to suggest that more effective data perturbation techniques could be proposed.

It is worth highlighting that LDP exhibits certain flaws in our experiments. Specifically, when large perturbations are applied, LDP results in significant performance degradation without effectively providing additional privacy protection. This raises important questions about the limitations of LDP, particularly when large perturbations are needed for strong privacy guarantees. Future research could explore whether LDP can be improved by adopting adaptive perturbation schemes or combining it with other privacy-preserving techniques to better balance privacy and performance.

\begin{table*}[t]
\caption{Performance of the FedEM Algorithm with Different Perturbation Iterations.}
\label{table-iter}
\vskip 0.15in
\begin{center}
\begin{small}
\begin{sc}
\begin{tabular}{lccccccc}
\toprule
\textbf{Dataset} & \textbf{Iterations} & \textbf{Test-Acc} & \textbf{Val-Acc} & \textbf{Test-MSE} & \textbf{Fea MSE} & \textbf{SSIM} & \textbf{PSNR} \\
\midrule
\multirow{3}{*}{MNIST} 
    & 1 & 0.9716 & 0.9672 & 1.2061 & 7.9397 & 0.1084 & 9.8520 \\
    & 5 & 0.9723 & 0.9684 & 1.1800 & 7.7600 & 0.1131 & 9.8471 \\
    & 10 & 0.9714 & 0.9667 & 1.3098 & 9.3221 & 0.0966 & 9.3797 \\
\midrule
\multirow{3}{*}{FMNIST} 
    & 1 & 0.8489 & 0.8455 & 0.9481 & 2.0721 & 0.1173 & 7.2210 \\
    & 5 & 0.8499 & 0.8452 & 0.8213 & 1.7213 & 0.1454 & 10.032 \\
    & 10 & 0.8512 & 0.8452 & 0.8087 & 1.5947 & 0.1740 & 10.465 \\
\midrule
\multirow{3}{*}{CIFAR-10} 
    & 1 & 0.2498 & 0.2393 & 2.4212 & 7.6672 & 0.0170 & 9.0659 \\
    & 5 & 0.2554 & 0.2407 & 1.7646 & 6.5318 & 0.0170 & 9.6219 \\
    & 10 & 0.2479 & 0.2325 & 2.4731 & 8.1993 & 0.0136 & 8.2652 \\
\bottomrule
\end{tabular}
\end{sc}
\end{small}
\end{center}
\vskip -0.1in
\end{table*}

\begin{figure*}[!htbp]
    \centering
    \includegraphics[width=\textwidth]{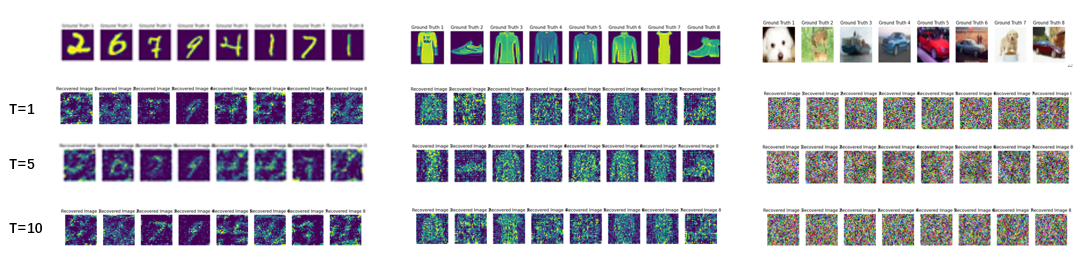} 
    \caption{DLG Reconstruction Images with Different Iteration Counts Under FedEM.}
    \label{fig:iter-res}
\end{figure*}

\subsection{Impact of Different Settings}

In this section, we further investigate the impact of parameter settings, such as the number of iterations used to generate data perturbations and the perturbation lower bound, on the defense strategy. Due to space limitations, we focus on discussing the effect of varying the number of iterations from 1 to 10 (with the default perturbation generation process involving five iterations). All other conditions remain consistent with those specified in Section~\ref{settings}. The impact of the perturbation lower bound on our method can be found in Section~\ref{lowerbound-full-results}.

The evaluation metrics for our FedEM method under different iteration counts are presented in Table~\ref{table-iter}, while the DLG attack-reconstructed images for various iterations are shown in Figure~\ref{fig:iter-res}(detailed in Appendix~\ref{iter-full-results}) . Analysis of the data and images indicates that, within a certain range, increasing the number of iterations improves model utility but degrades privacy protection. However, as the number of iterations grows, neither performance nor privacy exhibits a simple linear relationship with the iteration count, and this relationship warrants further exploration.

\begin{figure}[htbp]
    \centering
    \includegraphics[width=\linewidth]{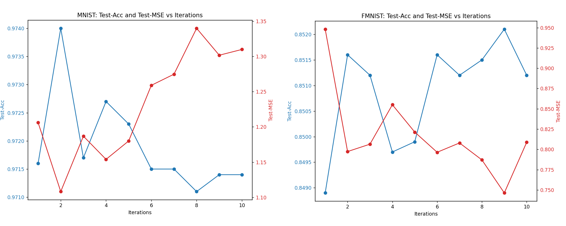} 
    \caption{Privacy-Utility Trade-off at Different Iteration Counts.}
    \label{fig:iter-trade-off}
\end{figure}

We now delve into the relationship between privacy and utility. To better understand this trade-off, we combine privacy and utility on the same graph. This combined analysis is crucial because it allows us to directly observe how changes in perturbation iteration count affect both privacy protection and model performance in tandem. Specifically, we plot the trade-off by setting the x-axis to the number of iterations used in the perturbation generation process, and the y-axis includes two metrics: Test-Acc, measuring model performance, and Test-MSE, indicating the level of privacy protection. The specific results are shown in Figure ~\ref{fig:iter-trade-off}.

From the observation of this figure, we conclude that when one metric (privacy or performance) reaches a high value, the other is inevitably compromised. Increasing privacy protection typically reduces model accuracy, while optimizing performance weakens privacy. This aligns with the "No Free Lunch" (NFL) theorem discussed in~\cite{zhang2022no}, and highlights an inherent challenge in federated learning systems—the conflict between performance and privacy.

\section{Conclusion}

In this paper, we introduced FedEM, a novel algorithm designed to defend against gradient leakage attacks in federated learning while preserving model utility. By integrating controlled perturbations into clients' local data, FedEM provides robust privacy protection without significantly compromising accuracy. Evaluations on datasets such as MNIST, FMNIST, and CIFAR-10 demonstrate that FedEM outperforms traditional privacy-preserving methods in both privacy and utility.

Future work will focus on enhancing perturbation strategies, exploring additional data perturbation techniques for federated learning, and integrating FedEM with other privacy-enhancing methods, such as homomorphic encryption. Furthermore, investigating the use of FedEM in dynamic and heterogeneous environments, such as healthcare and finance, opens up exciting possibilities for improving privacy protections in real-world applications. Moreover, future studies could also explore the performance of FedEM in the context of adversarial attacks beyond gradient leakage.

\section*{Impact Statement}

This paper presents FedEM, a method designed to enhance privacy protection in federated learning while maintaining model utility. The work contributes to advancing privacy-preserving techniques in machine learning, with potential applications in sensitive domains like healthcare and finance. There are no significant ethical concerns specific to this work, as it aligns with established privacy practices in federated learning.

\bibliography{example_paper}

\begin{thebibliography}{48}
\providecommand{\natexlab}[1]{#1}
\providecommand{\url}[1]{\texttt{#1}}
\expandafter\ifx\csname urlstyle\endcsname\relax
  \providecommand{\doi}[1]{doi: #1}\else
  \providecommand{\doi}{doi: \begingroup \urlstyle{rm}\Url}\fi

\bibitem[Aono et~al.(2017)Aono, Hayashi, Wang, Moriai, et~al.]{aono2017privacy}
Aono, Y., Hayashi, T., Wang, L., Moriai, S., et~al.
\newblock Privacy-preserving deep learning via additively homomorphic encryption.
\newblock \emph{IEEE transactions on information forensics and security}, 13\penalty0 (5):\penalty0 1333--1345, 2017.

\bibitem[Boenisch et~al.(2023)Boenisch, Dziedzic, Schuster, Shamsabadi, Shumailov, and Papernot]{boenisch2023curious}
Boenisch, F., Dziedzic, A., Schuster, R., Shamsabadi, A.~S., Shumailov, I., and Papernot, N.
\newblock When the curious abandon honesty: Federated learning is not private.
\newblock In \emph{2023 IEEE 8th European Symposium on Security and Privacy (EuroS\&P)}, pp.\  175--199. IEEE, 2023.

\bibitem[Bonawitz et~al.(2017)Bonawitz, Ivanov, Kreuter, Marcedone, McMahan, Patel, Ramage, Segal, and Seth]{bonawitz2017practical}
Bonawitz, K., Ivanov, V., Kreuter, B., Marcedone, A., McMahan, H.~B., Patel, S., Ramage, D., Segal, A., and Seth, K.
\newblock Practical secure aggregation for privacy-preserving machine learning.
\newblock In \emph{proceedings of the 2017 ACM SIGSAC Conference on Computer and Communications Security}, pp.\  1175--1191, 2017.

\bibitem[Chen \& Campbell(2021)Chen and Campbell]{chen2021understanding}
Chen, C. and Campbell, N.~D.
\newblock Understanding training-data leakage from gradients in neural networks for image classification.
\newblock \emph{arXiv preprint arXiv:2111.10178}, 2021.

\bibitem[Deng(2012)]{deng2012mnist}
Deng, L.
\newblock The mnist database of handwritten digit images for machine learning research [best of the web].
\newblock \emph{IEEE signal processing magazine}, 29\penalty0 (6):\penalty0 141--142, 2012.

\bibitem[Fowl et~al.(2021{\natexlab{a}})Fowl, Chiang, Goldblum, Geiping, Bansal, Czaja, and Goldstein]{fowl2021preventing}
Fowl, L., Chiang, P.-y., Goldblum, M., Geiping, J., Bansal, A., Czaja, W., and Goldstein, T.
\newblock Preventing unauthorized use of proprietary data: Poisoning for secure dataset release.
\newblock \emph{arXiv preprint arXiv:2103.02683}, 2021{\natexlab{a}}.

\bibitem[Fowl et~al.(2021{\natexlab{b}})Fowl, Geiping, Czaja, Goldblum, and Goldstein]{fowl2021robbing}
Fowl, L., Geiping, J., Czaja, W., Goldblum, M., and Goldstein, T.
\newblock Robbing the fed: Directly obtaining private data in federated learning with modified models.
\newblock \emph{arXiv preprint arXiv:2110.13057}, 2021{\natexlab{b}}.

\bibitem[Fowl et~al.(2022)Fowl, Geiping, Reich, Wen, Czaja, Goldblum, and Goldstein]{fowl2022decepticons}
Fowl, L., Geiping, J., Reich, S., Wen, Y., Czaja, W., Goldblum, M., and Goldstein, T.
\newblock Decepticons: Corrupted transformers breach privacy in federated learning for language models.
\newblock \emph{arXiv preprint arXiv:2201.12675}, 2022.

\bibitem[Fredrikson et~al.(2015)Fredrikson, Jha, and Ristenpart]{fredrikson2015model}
Fredrikson, M., Jha, S., and Ristenpart, T.
\newblock Model inversion attacks that exploit confidence information and basic countermeasures.
\newblock In \emph{Proceedings of the 22nd ACM SIGSAC conference on computer and communications security}, pp.\  1322--1333, 2015.

\bibitem[Ganju et~al.(2018)Ganju, Wang, Yang, Gunter, and Borisov]{ganju2018property}
Ganju, K., Wang, Q., Yang, W., Gunter, C.~A., and Borisov, N.
\newblock Property inference attacks on fully connected neural networks using permutation invariant representations.
\newblock In \emph{Proceedings of the 2018 ACM SIGSAC conference on computer and communications security}, pp.\  619--633, 2018.

\bibitem[Geiping et~al.(2020)Geiping, Bauermeister, Dr{\"o}ge, and Moeller]{geiping2020inverting}
Geiping, J., Bauermeister, H., Dr{\"o}ge, H., and Moeller, M.
\newblock Inverting gradients-how easy is it to break privacy in federated learning?
\newblock \emph{Advances in neural information processing systems}, 33:\penalty0 16937--16947, 2020.

\bibitem[Geyer et~al.(2017)Geyer, Klein, and Nabi]{geyer2017differentially}
Geyer, R.~C., Klein, T., and Nabi, M.
\newblock Differentially private federated learning: A client level perspective.
\newblock \emph{arXiv preprint arXiv:1712.07557}, 2017.

\bibitem[Girgis et~al.(2021)Girgis, Data, Diggavi, Kairouz, and Suresh]{girgis2021shuffled}
Girgis, A., Data, D., Diggavi, S., Kairouz, P., and Suresh, A.~T.
\newblock Shuffled model of differential privacy in federated learning.
\newblock In \emph{International Conference on Artificial Intelligence and Statistics}, pp.\  2521--2529. PMLR, 2021.

\bibitem[Huang et~al.(2021)Huang, Ma, Erfani, Bailey, and Wang]{huang2021unlearnable}
Huang, H., Ma, X., Erfani, S.~M., Bailey, J., and Wang, Y.
\newblock Unlearnable examples: Making personal data unexploitable.
\newblock \emph{arXiv preprint arXiv:2101.04898}, 2021.

\bibitem[Hui et~al.(2021)Hui, Yang, Yuan, Burlina, Gong, and Cao]{hui2021practical}
Hui, B., Yang, Y., Yuan, H., Burlina, P., Gong, N.~Z., and Cao, Y.
\newblock Practical blind membership inference attack via differential comparisons.
\newblock \emph{arXiv preprint arXiv:2101.01341}, 2021.

\bibitem[Jiang et~al.(2024)Jiang, Wang, Que, and Lin]{jiang2024fed}
Jiang, S., Wang, X., Que, Y., and Lin, H.
\newblock Fed-mps: Federated learning with local differential privacy using model parameter selection for resource-constrained cps.
\newblock \emph{Journal of Systems Architecture}, 150:\penalty0 103108, 2024.

\bibitem[Kim et~al.(2021)Kim, G{\"u}nl{\"u}, and Schaefer]{kim2021federated}
Kim, M., G{\"u}nl{\"u}, O., and Schaefer, R.~F.
\newblock Federated learning with local differential privacy: Trade-offs between privacy, utility, and communication.
\newblock In \emph{ICASSP 2021-2021 IEEE International Conference on Acoustics, Speech and Signal Processing (ICASSP)}, pp.\  2650--2654. IEEE, 2021.

\bibitem[Krizhevsky et~al.(2009)Krizhevsky, Hinton, et~al.]{krizhevsky2009learning}
Krizhevsky, A., Hinton, G., et~al.
\newblock Learning multiple layers of features from tiny images.
\newblock 2009.

\bibitem[Li et~al.(2024)Li, Chen, and Teng]{li2024comprehensive}
Li, J., Chen, T., and Teng, S.
\newblock A comprehensive survey on client selection strategies in federated learning.
\newblock \emph{Computer Networks}, pp.\  110663, 2024.

\bibitem[Liu et~al.(2020)Liu, Cao, Yoshikawa, and Chen]{liu2020fedsel}
Liu, R., Cao, Y., Yoshikawa, M., and Chen, H.
\newblock Fedsel: Federated sgd under local differential privacy with top-k dimension selection.
\newblock In \emph{Database Systems for Advanced Applications: 25th International Conference, DASFAA 2020, Jeju, South Korea, September 24--27, 2020, Proceedings, Part I 25}, pp.\  485--501. Springer, 2020.

\bibitem[Lu et~al.(2022)Lu, Zhang, Zhao, He, and Cheng]{lu2022april}
Lu, J., Zhang, X.~S., Zhao, T., He, X., and Cheng, J.
\newblock April: Finding the achilles' heel on privacy for vision transformers.
\newblock In \emph{Proceedings of the IEEE/CVF Conference on Computer Vision and Pattern Recognition}, pp.\  10051--10060, 2022.

\bibitem[Madi et~al.(2021)Madi, Stan, Mayoue, Grivet-S{\'e}bert, Gouy-Pailler, and Sirdey]{madi2021secure}
Madi, A., Stan, O., Mayoue, A., Grivet-S{\'e}bert, A., Gouy-Pailler, C., and Sirdey, R.
\newblock A secure federated learning framework using homomorphic encryption and verifiable computing.
\newblock In \emph{2021 Reconciling Data Analytics, Automation, Privacy, and Security: A Big Data Challenge (RDAAPS)}, pp.\  1--8. IEEE, 2021.

\bibitem[McMahan et~al.(2017)McMahan, Moore, Ramage, Hampson, and y~Arcas]{mcmahan2017communication}
McMahan, B., Moore, E., Ramage, D., Hampson, S., and y~Arcas, B.~A.
\newblock Communication-efficient learning of deep networks from decentralized data.
\newblock In \emph{Artificial intelligence and statistics}, pp.\  1273--1282. PMLR, 2017.

\bibitem[Melis et~al.(2019)Melis, Song, De~Cristofaro, and Shmatikov]{melis2019exploiting}
Melis, L., Song, C., De~Cristofaro, E., and Shmatikov, V.
\newblock Exploiting unintended feature leakage in collaborative learning.
\newblock In \emph{2019 IEEE symposium on security and privacy (SP)}, pp.\  691--706. IEEE, 2019.

\bibitem[Miao et~al.(2022)Miao, Yang, Hu, Li, and Huang]{miao2022against}
Miao, L., Yang, W., Hu, R., Li, L., and Huang, L.
\newblock Against backdoor attacks in federated learning with differential privacy.
\newblock In \emph{ICASSP 2022-2022 IEEE International Conference on Acoustics, Speech and Signal Processing (ICASSP)}, pp.\  2999--3003. IEEE, 2022.

\bibitem[Nasr et~al.(2019)Nasr, Shokri, and Houmansadr]{nasr2019comprehensive}
Nasr, M., Shokri, R., and Houmansadr, A.
\newblock Comprehensive privacy analysis of deep learning: Passive and active white-box inference attacks against centralized and federated learning.
\newblock In \emph{2019 IEEE symposium on security and privacy (SP)}, pp.\  739--753. IEEE, 2019.

\bibitem[Salem et~al.(2018)Salem, Zhang, Humbert, Berrang, Fritz, and Backes]{salem2018ml}
Salem, A., Zhang, Y., Humbert, M., Berrang, P., Fritz, M., and Backes, M.
\newblock Ml-leaks: Model and data independent membership inference attacks and defenses on machine learning models.
\newblock \emph{arXiv preprint arXiv:1806.01246}, 2018.

\bibitem[Shokri et~al.(2017)Shokri, Stronati, Song, and Shmatikov]{shokri2017membership}
Shokri, R., Stronati, M., Song, C., and Shmatikov, V.
\newblock Membership inference attacks against machine learning models.
\newblock In \emph{2017 IEEE symposium on security and privacy (SP)}, pp.\  3--18. IEEE, 2017.

\bibitem[Song \& Mittal(2021)Song and Mittal]{song2021systematic}
Song, L. and Mittal, P.
\newblock Systematic evaluation of privacy risks of machine learning models.
\newblock In \emph{30th USENIX Security Symposium (USENIX Security 21)}, pp.\  2615--2632, 2021.

\bibitem[Sun et~al.(2020)Sun, Qian, and Chen]{sun2020ldp}
Sun, L., Qian, J., and Chen, X.
\newblock Ldp-fl: Practical private aggregation in federated learning with local differential privacy.
\newblock \emph{arXiv preprint arXiv:2007.15789}, 2020.

\bibitem[Wang et~al.(2023{\natexlab{a}})Wang, Chen, Jiang, and Zhao]{wang2023ppefl}
Wang, B., Chen, Y., Jiang, H., and Zhao, Z.
\newblock Ppefl: Privacy-preserving edge federated learning with local differential privacy.
\newblock \emph{IEEE Internet of Things Journal}, 10\penalty0 (17):\penalty0 15488--15500, 2023{\natexlab{a}}.

\bibitem[Wang et~al.(2004)Wang, Bovik, Sheikh, and Simoncelli]{wang2004image}
Wang, Z., Bovik, A.~C., Sheikh, H.~R., and Simoncelli, E.~P.
\newblock Image quality assessment: from error visibility to structural similarity.
\newblock \emph{IEEE transactions on image processing}, 13\penalty0 (4):\penalty0 600--612, 2004.

\bibitem[Wang et~al.(2023{\natexlab{b}})Wang, Liu, Hu, Ren, Guo, and Yuan]{wang2023attrleaks}
Wang, Z., Liu, K., Hu, J., Ren, J., Guo, H., and Yuan, W.
\newblock Attrleaks on the edge: Exploiting information leakage from privacy-preserving co-inference.
\newblock \emph{Chinese Journal of Electronics}, 32\penalty0 (1):\penalty0 1--12, 2023{\natexlab{b}}.

\bibitem[Xiao et~al.(2017)Xiao, Rasul, and Vollgraf]{xiao2017fashion}
Xiao, H., Rasul, K., and Vollgraf, R.
\newblock Fashion-mnist: a novel image dataset for benchmarking machine learning algorithms.
\newblock \emph{arXiv preprint arXiv:1708.07747}, 2017.

\bibitem[Xu et~al.(2019)Xu, Li, Liu, Yang, and Lin]{xu2019verifynet}
Xu, G., Li, H., Liu, S., Yang, K., and Lin, X.
\newblock Verifynet: Secure and verifiable federated learning.
\newblock \emph{IEEE Transactions on Information Forensics and Security}, 15:\penalty0 911--926, 2019.

\bibitem[Yin et~al.(2021)Yin, Mallya, Vahdat, Alvarez, Kautz, and Molchanov]{yin2021see}
Yin, H., Mallya, A., Vahdat, A., Alvarez, J.~M., Kautz, J., and Molchanov, P.
\newblock See through gradients: Image batch recovery via gradinversion.
\newblock In \emph{Proceedings of the IEEE/CVF conference on computer vision and pattern recognition}, pp.\  16337--16346, 2021.

\bibitem[Yue et~al.(2023)Yue, Jin, Wong, Baron, and Dai]{yue2023gradient}
Yue, K., Jin, R., Wong, C.-W., Baron, D., and Dai, H.
\newblock Gradient obfuscation gives a false sense of security in federated learning.
\newblock In \emph{32nd USENIX Security Symposium (USENIX Security 23)}, pp.\  6381--6398, 2023.

\bibitem[Zhang et~al.(2022)Zhang, Gu, Fan, Chen, and Yang]{zhang2022no}
Zhang, X., Gu, H., Fan, L., Chen, K., and Yang, Q.
\newblock No free lunch theorem for security and utility in federated learning.
\newblock \emph{ACM Transactions on Intelligent Systems and Technology}, 14\penalty0 (1):\penalty0 1--35, 2022.

\bibitem[Zhang et~al.(2023{\natexlab{a}})Zhang, Fan, Wang, Li, Chen, and Yang]{zhang2023game}
Zhang, X., Fan, L., Wang, S., Li, W., Chen, K., and Yang, Q.
\newblock A game-theoretic framework for federated learning.
\newblock \emph{arXiv preprint arXiv:2304.05836}, 2023{\natexlab{a}}.

\bibitem[Zhang et~al.(2023{\natexlab{b}})Zhang, Huang, Fan, Chen, and Yang]{zhang2023probably}
Zhang, X., Huang, A., Fan, L., Chen, K., and Yang, Q.
\newblock Probably approximately correct federated learning.
\newblock \emph{arXiv preprint arXiv:2304.04641}, 2023{\natexlab{b}}.

\bibitem[Zhang et~al.(2023{\natexlab{c}})Zhang, Li, Chen, Xia, and Yang]{zhang2023theoretically}
Zhang, X., Li, W., Chen, K., Xia, S., and Yang, Q.
\newblock Theoretically principled federated learning for balancing privacy and utility.
\newblock \emph{arXiv preprint arXiv:2305.15148}, 2023{\natexlab{c}}.

\bibitem[Zhang et~al.(2024)Zhang, Kang, Fan, Chen, and Yang]{zhang2024meta}
Zhang, X., Kang, Y., Fan, L., Chen, K., and Yang, Q.
\newblock A meta-learning framework for tuning parameters of protection mechanisms in trustworthy federated learning.
\newblock \emph{ACM Transactions on Intelligent Systems and Technology}, 15\penalty0 (3):\penalty0 1--36, 2024.

\bibitem[Zhao et~al.(2020{\natexlab{a}})Zhao, Mopuri, and Bilen]{zhao2020idlg}
Zhao, B., Mopuri, K.~R., and Bilen, H.
\newblock idlg: Improved deep leakage from gradients.
\newblock \emph{arXiv preprint arXiv:2001.02610}, 2020{\natexlab{a}}.

\bibitem[Zhao et~al.(2022)Zhao, Zhu, Wang, Lu, Liu, and Li]{zhao2022pvd}
Zhao, J., Zhu, H., Wang, F., Lu, R., Liu, Z., and Li, H.
\newblock Pvd-fl: A privacy-preserving and verifiable decentralized federated learning framework.
\newblock \emph{IEEE Transactions on Information Forensics and Security}, 17:\penalty0 2059--2073, 2022.

\bibitem[Zhao et~al.(2020{\natexlab{b}})Zhao, Zhao, Yang, Wang, Wang, Lyu, Niyato, and Lam]{zhao2020local}
Zhao, Y., Zhao, J., Yang, M., Wang, T., Wang, N., Lyu, L., Niyato, D., and Lam, K.-Y.
\newblock Local differential privacy-based federated learning for internet of things.
\newblock \emph{IEEE Internet of Things Journal}, 8\penalty0 (11):\penalty0 8836--8853, 2020{\natexlab{b}}.

\bibitem[Zheng et~al.(2020)Zheng, Zeng, Zhou, Hsieh, Cheng, and Huang]{zheng2020evaluating}
Zheng, X., Zeng, J., Zhou, Y., Hsieh, C.-J., Cheng, M., and Huang, X.-J.
\newblock Evaluating and enhancing the robustness of neural network-based dependency parsing models with adversarial examples.
\newblock In \emph{Proceedings of the 58th Annual Meeting of the Association for Computational Linguistics}, pp.\  6600--6610, 2020.

\bibitem[Zhu \& Blaschko(2020)Zhu and Blaschko]{zhu2020r}
Zhu, J. and Blaschko, M.
\newblock R-gap: Recursive gradient attack on privacy.
\newblock \emph{arXiv preprint arXiv:2010.07733}, 2020.

\bibitem[Zhu et~al.(2019)Zhu, Liu, and Han]{zhu2019deep}
Zhu, L., Liu, Z., and Han, S.
\newblock Deep leakage from gradients.
\newblock \emph{Advances in neural information processing systems}, 32, 2019.

\end{thebibliography}
\bibliographystyle{icml2025}

\newpage
\appendix
\onecolumn

\section{Notations used in this article.}\label{APP A}

\begin{table}[ht]
\centering
\caption{Notation for Related Works and FedEM Algorithm}
\begin{tabular}{lccr}
\hline\hline
\textbf{Symbol} & \textbf{Description} \\
\hline
$K$ & Total number of clients in federated learning \\
\hline
$D_k$ & Local dataset held by client $k$ \\
\hline
$m_k$ & Number of data points available to client $k$ \\
\hline
$m$ & Total number of data points across all clients, $m = \sum_{k=1}^{K} m_k$ \\
\hline
$\theta$ & Global model parameters \\
\hline
$\theta_u$ & Local model parameters for client $k$ during perturbation update \\
\hline
$L$ & Loss function \\
\hline
$\mathcal{L}_k$ & Loss function for client $k$ \\
\hline
$g_k$ & Local gradient computed by client $k$ \\
\hline
$\delta_k$ & Perturbation vector for client $k$ \\
\hline
$f_\theta$ & Predictive model with parameters $\theta$ \\
\hline
$t(x_k + \delta_k)$ & Transformation or preprocessing applied to perturbed input \\
\hline
$\mathbb{E}$ & Expectation \\
\hline
$N$ & Number of perturbation iterations \\
\hline
$\alpha_u$ & Step size for perturbation update \\
\hline
$\rho_u^{\text{max}}$ & Maximum allowed norm for perturbation $\delta_k$ \\
\hline
$\rho_u^{\text{min}}$ & Minimum allowed norm for perturbation $\delta_k$ \\
\hline
$\eta$ & Learning rate for model parameter update \\
\hline
$g_{\text{global}}$ & Aggregated gradient computed by the server from all clients \\
\hline
\end{tabular}
\end{table}

\section{Detailed description for the experimental setup.}\label{APP B}

\subsection{Datasets}

MNIST is a grayscale image classification dataset consisting of 10 digit classes. It includes 60,000 training images and 10,000 test images. FashionMNIST is a grayscale image dataset similar to MNIST, but with 10 clothing categories. It consists of 60,000 training images and 10,000 test images. CIFAR-10 is a color image classification dataset with 10 daily object categories. It contains 50,000 training samples and 10,000 test samples.

\subsection{Attack Method}

We choose DLG\cite{zhu2019deep} attack as the threat model, which is an attack targeting privacy leakage in federated learning, aiming to reconstruct the original training data samples from shared gradient updates. The core idea of the attack is to minimize the discrepancy between the gradient computed from reconstructed data and the shared gradients, thus recovering data points that closely resemble the original ones. The optimization objective for DLG is:

\begin{equation}
\min_{{x}}\min_{{y}} \|\nabla_{\theta}\mathcal{L}(x, y) - g\|
\end{equation}

where $\nabla_{\theta}\mathcal{L}(x, y)$ represents the gradient computed using the reconstructed input $x$ and label $y$, and $g$ represents the true shared gradient. The goal of the attack is to minimize the difference between the gradients of the reconstructed data and the shared gradients, thus reconstructing data samples that resemble the originals as closely as possible.

The pseudocode for the DLG attack is as follows:

\begin{algorithm}[H]
\caption{DLG Attack}
\begin{algorithmic}[1]
\STATE \textbf{Input:} Shared gradient $g$, loss function $\mathcal{L}$, initial input $x_{orig}$, learning rate $\eta$, maximum number of iterations $T$
\STATE Initialize $x \leftarrow x_{orig}$, $y \leftarrow \text{random label}$
\FOR{t = 1 to T}
    \STATE Compute gradient $\nabla_{\theta}\mathcal{L}(x, y)$ for the current $x$ and $y$
    \STATE Compute loss $L = \|\nabla_{\theta}\mathcal{L}(x, y) - g\|_2$
    \STATE Update $x \leftarrow x - \eta \cdot \nabla_x L$
    \STATE Update $y \leftarrow y - \eta \cdot \nabla_y L$
\ENDFOR
\STATE \textbf{Output:} Reconstructed data $x$
\end{algorithmic}
\end{algorithm}

\subsection{LDP Method}

LDP is a privacy-preserving technique. In the context of federated learning, the LDP method involves adding noise to the gradients calculated by each client. This noise is typically drawn from a distribution such as the Laplace or Gaussian distribution. The goal is to prevent any single client's data from being distinguishable from others, ensuring that an adversary cannot infer the presence of any specific data point based on the gradients.

The optimization objective in LDP for federated learning is similar to that in traditional federated learning, but with the addition of noise to the gradients before they are sent to the server. The objective is to minimize the loss function while ensuring that the gradients remain differentially private. The LDP optimization problem can be expressed as:
\begin{equation}
\min_{\theta} \sum_{k=1}^{K} \lambda_k \mathbb{E}_{(x_k, y_k) \sim D_k}[\mathcal{L}(f(x_k; \theta), y_k)],
\end{equation}
where $x_k$ is the local data of client $k$, $y_k$ is the corresponding label, and $\lambda_k$ is the weight associated with each client. 

The following pseudocode optimizes the LDP method for a multi-round federated learning process:
\begin{algorithm}[H]
\caption{LDP Method for Federated Learning}
\begin{algorithmic}[1]
\STATE \textbf{Input:} Local data $T_k$ for each client, model $\mathcal{L}$, noise distribution $P(\delta)$, learning rate $\eta$, number of rounds $R$, number of clients $K$
\STATE Initialize global model parameters $\theta$
\FOR{round $r = 1$ to $R$}
    \STATE \textbf{Client Update:}
    \FOR{each client $k$ in parallel}
        \STATE Compute gradient $\nabla_{\theta_k} \mathcal{L}(x_k, y_k)$ for client $k$
        \STATE Add noise $\delta_k \sim P(\delta)$ to the gradient
        \STATE Send noisy gradient to the server
    \ENDFOR
    \STATE \textbf{Server Aggregation:}
    \STATE Aggregate gradients from all clients: $\nabla_{\theta} = \sum_{k=1}^{K} \lambda_k \nabla_{\theta_k}$, where $\lambda_k$ is the weight of client $k$
    \STATE Update global model parameters: $\theta \leftarrow \theta - \eta \nabla_{\theta}$
\ENDFOR
\STATE \textbf{Output:} Final global model parameters $\theta$
\end{algorithmic}
\end{algorithm}

\subsection{Evaluation Metrics}

To assess the privacy protection of the methods used in this study, we employ three standard evaluation metrics: mean squared error (MSE), structural similarity index measure (SSIM)~\cite{wang2004image}, and peak signal-to-noise ratio (PSNR). These metrics are used to compare the reconstructed images with the original ones, providing insight into the effectiveness of the privacy protection techniques.

MSE measures the average squared difference between the original and reconstructed images. A lower MSE indicates that the reconstructed image is closer to the original, implying less privacy protection. Therefore, higher MSE values are preferred, as they suggest greater obfuscation of the original data, which is indicative of stronger privacy protection.

SSIM evaluates the perceptual similarity between the original and reconstructed images, taking into account luminance, contrast, and structural information. A higher SSIM value indicates better preservation of the original image structure, which would suggest weaker privacy protection. Hence, lower SSIM values are desirable, as they reflect more distortion and stronger privacy protection.

PSNR measures the quality of the reconstructed image relative to the original, with higher PSNR values indicating better quality and, therefore, less privacy protection. As with MSE, higher PSNR values suggest weaker privacy protection.

\section{Additional experimental results.}\label{APP C}

\begin{table*}[t]
\caption{Changes in Evaluation Metrics for different lower bounds of perturbation.($\rho_{\text{min}}$ and $\rho_{\text{max}}$ are the bounds of the perturbation norm.)}
\label{table-lower-bound}
\vskip 0.15in
\begin{center}
\begin{small}
\begin{sc}
\begin{tabular}{lccccccc}
\toprule
\textbf{Dataset} & \textbf{Iterations} & \textbf{Test-Acc} & \textbf{Val-Acc} & \textbf{Test-MSE} & \textbf{Fea MSE} & \textbf{SSIM} & \textbf{PSNR} \\
\midrule
\multirow{3}{*}{MNIST} 
    & 0 & 0.9801 & 0.9759 & 1.1972 & 1.6294 & 0.1299 & 9.7649 \\
    & \(\rho_{\text{min}} = \frac{1}{8} \rho_{\text{max}}\) & 0.9787 & 0.9725 & 1.1402 & 1.8706 & 0.1116 & 10.319 \\
    & \(\rho_{\text{min}} = \frac{1}{4} \rho_{\text{max}}\) & 0.9779 & 0.9793 & 1.2001 & 5.8275 & 0.1241 & 10.062 \\
\midrule
\multirow{3}{*}{FMNIST} 
    & 0 & 0.8605 & 0.8398 & 0.8439 & 2.2711 & 0.2225 & 10.167 \\
    & \(\rho_{\text{min}} = \frac{1}{8} \rho_{\text{max}}\) & 0.8563 & 0.8507 & 0.9002 & 2.0153 & 0.2345 & 10.186 \\
    & \(\rho_{\text{min}} = \frac{1}{4} \rho_{\text{max}}\) & 0.8515 & 0.8531 & 1.1095 & 10.584 & 0.2325 & 9.4053 \\
\midrule
\multirow{3}{*}{CIFAR-10} 
    & 0 & 0.2829 & 0.2953 & 1.8536 & 7.2957 & 0.0153 & 9.4162 \\
    & \(\rho_{\text{min}} = \frac{1}{8} \rho_{\text{max}}\) & 0.2782 & 0.2675 & 1.5687 & 10.775 & 0.0159 & 10.142 \\
    & \(\rho_{\text{min}} = \frac{1}{4} \rho_{\text{max}}\) & 0.2709 & 0.2649 & 2.0489 & 14.094 & 0.0167 & 9.0745 \\
\bottomrule
\end{tabular}
\end{sc}
\end{small}
\end{center}
\vskip -0.1in
\end{table*}

\begin{figure*}[!htbp]
    \centering
    \includegraphics[width=\textwidth]{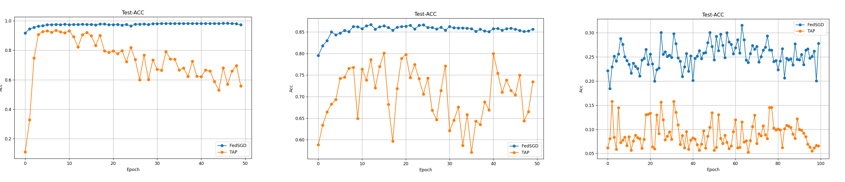} 
    \caption{Variation in Test Set Accuracy for FedSGD with and without TAP Perturbation. From left to right: MNIST, FMNIST, and CIFAR-10 datasets.}
    \label{fig:TAP}
\end{figure*}

\subsection{The performance of FedEM under different lower bounds of perturbation.}\label{lowerbound-full-results}

In this subsection, we investigate the impact of the perturbation lower bound on FedEM by varying its size. The specific experimental results are shown in Table~\ref{lowerbound-full-results} (Note that, to make the differences caused by varying perturbation sizes more apparent, no data normalization was performed in this set of experiments). From the data in the table, it can be observed that the accuracy of both the test and training sets exhibits a certain linear relationship with the perturbation size. As the perturbation increases, the accuracy decreases accordingly. Additionally, by observing the privacy protection-related metrics, it can be concluded that, to some extent, larger perturbations lead to better privacy protection outcomes.

\subsection{Detailed Experimental Results of Different Perturbation Generation Iterations}\label{iter-full-results}

\begin{figure}[htbp]
    \centering
    \includegraphics[width=\textwidth]{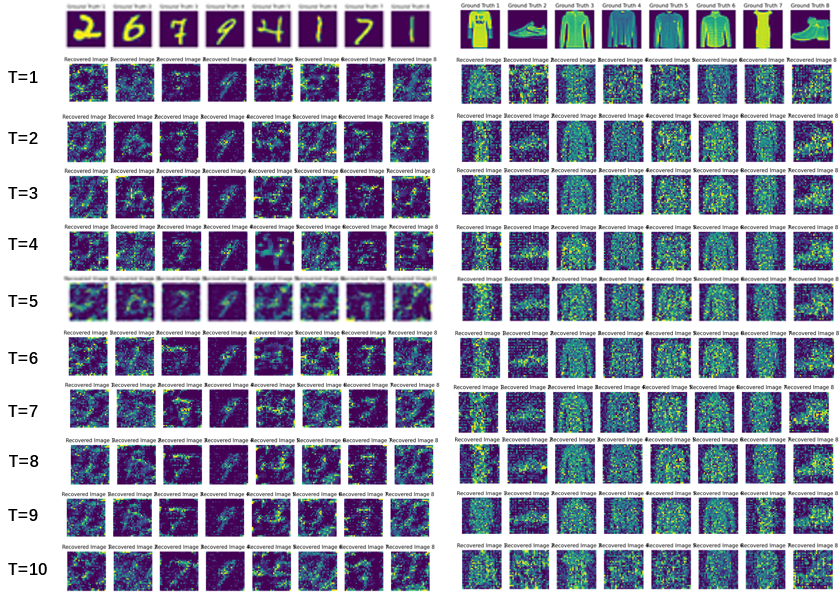} 
    \caption{DLG Reconstruction Images for Iteration Counts 1 to 10 Under the FedEM Framework.}
    \label{fig:full-iter-res}
\end{figure}

In this section, we present detailed results of FedEM on MNIST and FMNIST, where perturbations are generated using different iteration counts. The evaluation metrics for each iteration can be found in Table \ref{full-iter-tab}, and the comparison of reconstructed images across iterations is shown in Figure \ref{fig:full-iter-res}. As the accuracy changes minimally across different iterations, we have not included the accuracy curves in the paper.

\begin{table*}[htbp]
\caption{Changes in Different Evaluation Metrics for Perturbation Iterations from 1 to 10.}
\label{full-iter-tab}
\vskip 0.15in
\begin{center}
\begin{small}
\begin{sc}
\begin{tabular}{lccccccc}
\toprule
\textbf{Dataset} & \textbf{Iterations} & \textbf{Test-Acc} & \textbf{Val-Acc} & \textbf{Test-MSE} & \textbf{Fea MSE} & \textbf{SSIM} & \textbf{PSNR} \\
\midrule
\multirow{10}{*}{MNIST} 
   & 1 & 0.9716 & 0.9672 & 1.2061 & 7.9397 & 0.1084 & 9.8520 \\ 
   & 2 & 0.9740 & 0.9692 & 1.1082 & 5.9884 & 0.1186 & 10.170 \\ 
   & 3 & 0.9717 & 0.9669 & 1.1868 & 6.2519 & 0.1144 & 9.8295 \\ 
   & 4 & 0.9727 & 0.9685 & 1.1539 & 6.6599 & 0.1338 & 10.118 \\ 
   & 5 & 0.9723 & 0.9684 & 1.1800 & 7.7600 & 0.1131 & 9.8471 \\
   & 6 & 0.9715 & 0.9679 & 1.2589 & 7.9635 & 0.1076 & 9.6079 \\ 
   & 7 & 0.9715 & 0.9671 & 1.2747 & 8.5796 & 0.1110 & 9.6692 \\ 
   & 8 & 0.9711 & 0.9666 & 1.3398 & 7.0543 & 0.0819 & 9.2652 \\ 
   & 9 & 0.9714 & 0.9661 & 1.3016 & 7.4549 & 0.0964 & 9.4314 \\ 
   & 10 & 0.9714 & 0.9667 & 1.3098 & 9.3221 & 0.0966 & 9.3797 \\ 
\midrule
\multirow{10}{*}{FMNIST} 
   & 1 & 0.8498 & 0.8455 & 0.9481 & 2.0721 & 0.1173 & 9.7247 \\ 
   & 2 & 0.8516 & 0.8490 & 0.7973 & 1.7360 & 0.1682 & 10.228 \\ 
   & 3 & 0.8512 & 0.8482 & 0.8064 & 1.7059 & 0.1669 & 10.228 \\ 
   & 4 & 0.8497 & 0.8479 & 0.8550 & 2.5371 & 0.1488 & 9.9643 \\ 
   & 5 & 0.8499 & 0.8452 & 0.8213 & 1.7213 & 0.1454 & 10.032 \\ 
   & 6 & 0.8516 & 0.8459 & 0.7964 & 1.7963 & 0.1774 & 10.258 \\ 
   & 7 & 0.8512 & 0.8458 & 0.8079 & 1.6036 & 0.1670 & 10.169 \\ 
   & 8 & 0.8515 & 0.8470 & 0.7870 & 1.6466 & 0.1741 & 10.267 \\ 
   & 9 & 0.8521 & 0.8471 & 0.7463 & 0.8768 & 0.1740 & 10.465 \\ 
   & 10 & 0.8512 & 0.8452 & 0.8087 & 1.5947 & 0.1714 & 10.174 \\ 
\bottomrule
\end{tabular}
\end{sc}
\end{small}
\end{center}
\vskip -0.1in
\end{table*}

\subsection{Other Explorations of Data Distortion for Privacy Protection}

\begin{table*}[t]
\caption{Comparison of FedSGD and TAP Perturbation Results.}
\label{TAP-tab}
\vskip 0.15in
\begin{center}
\begin{small}
\begin{sc}
\begin{tabular}{lccccccc}
\toprule
\textbf{Dataset} & \textbf{Methods} & \textbf{Test-Acc} & \textbf{Val-Acc} & \textbf{Test-MSE} & \textbf{Fea MSE} & \textbf{SSIM} & \textbf{PSNR} \\
\midrule
\multirow{2}{*}{MNIST} 
   & FedSGD & 0.9784 & 0.9754 & 0.9324 & 2.0676 & 0.1573 & 11.387 \\
   & TAP & 0.9354 & 0.9360 & 0.8398 & 3.8185 & 0.3611 & 10.992 \\  
\midrule
\multirow{2}{*}{FMNIST} 
   & FedSGD & 0.8622 & 0.8542 & 0.8064 & 0.8206 & 0.2453 & 10.588 \\
   & TAP & 0.7999 & 0.7910 & 1.2741 & 14.486 & 0.0988 & 8.4439 \\ 
\midrule
\multirow{2}{*}{CIFAR-10}
    & FedSGD & 0.3003 & 0.2945 & 1.9111 & 14.366 & 0.0175 & 9.3946 \\
    & TAP & 0.1454 & 0.1375 & 1.1721 & 93.018 & 0.0174 & 11.809 \\
\bottomrule
\end{tabular}
\end{sc}
\end{small}
\end{center}
\vskip -0.1in
\end{table*}

We also explored other data perturbation algorithms. In this section, we present the experimental results of combining TAP\cite{fowl2021preventing} with federated learning. The comparison with the FedSGD algorithm, which does not include privacy protection, is shown in Table ~\ref{TAP-tab}, and the accuracy variation curve is depicted in Figure~\ref{fig:TAP}. From the data in the table, it is evident that TAP provides strong privacy protection. However, the training process does not converge, and thus, we have not included this method in the main body of the paper. We view this method as an indication that better data perturbation techniques are yet to be discovered.


\end{document}